%% file: main.tex
\documentclass{article}
\usepackage{spconf}
\usepackage{amsmath}
\usepackage{graphicx}
\usepackage{cite}

\graphicspath{{fig/}}
\usepackage[font=small,skip=5pt]{caption}
\usepackage{booktabs}
\usepackage{tabularx}
\usepackage{multirow}
\usepackage{bm}
\usepackage{xcolor}
\usepackage{microtype}

\title{Acoustically grounded word embeddings \\ for improved acoustics-to-word speech recognition}

\name{
    Shane Settle$^{\star}$ \qquad
    Kartik Audhkhasi$^{\dagger}$ \qquad
    Karen Livescu$^{\star}$ \qquad
    Michael Picheny$^{\dagger}$   
}
\address{
    $^{\star}$ TTI-Chicago \qquad
    $^{\dagger}$IBM Research AI
}

\usepackage{textpos}

\newcommand{\copyrightstatement}{
    \begin{textblock}{11.25}(1,10.5)
         \center
         \footnotesize \copyright 2019 IEEE.  Personal use of this material is permitted.  Permission from IEEE must be obtained for all other uses, in any current or future media, including reprinting/republishing this material for advertising or promotional purposes, creating new collective works, for resale or redistribution to servers or lists, or reuse of any copyrighted component of this work in other works.
    \end{textblock}
}

\begin{document}
\ninept

\maketitle

\copyrightstatement
\vspace{-.25in}
\input{abstract}
\input{intro}
\vspace{-.1in}
\input{approach}

\vspace{-.1in}
\input{experiments}

\vspace{-.2in}
\input{results}
\vspace{-.15in}
\input{conclusion}

\vfill\clearpage

\bibliographystyle{IEEEbib.bst}
\bibliography{refs}

\end{document}

%% file: abstract.tex
\begin{abstract}
  Direct acoustics-to-word (A2W) systems for end-to-end automatic speech recognition are simpler to train, and more efficient to decode with, than sub-word systems.
  However, A2W systems can have difficulties at training time when data is limited, and at decoding time when recognizing words outside the training vocabulary.
  To address these shortcomings, we investigate the use of recently proposed acoustic and acoustically grounded word embedding techniques in A2W systems.
  The idea is based on treating the final pre-softmax weight matrix of an AWE recognizer as a matrix of word embedding vectors, and using an externally trained set of word embeddings to improve the quality of this matrix.
  In particular we introduce two ideas:  (1) Enforcing similarity at training time between the external embeddings and the recognizer weights, and (2) using the word embeddings at test time for predicting out-of-vocabulary words.
  Our word embedding model is {\it acoustically grounded}, that is it is learned jointly with acoustic embeddings so as to encode the words' acoustic-phonetic content; and it is {\it parametric}, so that it can embed any arbitrary (potentially out-of-vocabulary) sequence of characters.
  We find that both techniques improve the performance of an A2W recognizer on conversational telephone speech.
\end{abstract}


%% file: intro.tex
\section{Introduction}
\label{sec:intro}

End-to-end automatic speech recognition (ASR) focuses on replacing the modular training approaches of traditional ASR systems with conceptually simpler methods. Instead of requiring separately trained acoustic, pronunciation, and language models, neural network-based connectionist temporal classification (CTC) and encoder-decoder
approaches allow for joint optimization of a single objective.  In principle such models can map acoustics directly to words. However, to achieve performance comparable with traditional methods, these systems are still typically trained to predict sub-word units such as characters or ``wordpieces''~\cite{rao2017exploring,chiu2018state,sanabria2018hierarchical,krishna2018hierarchical}, thereby relying on additional decoders and externally trained language models.

Acoustics-to-word (A2W) systems~\cite{soltau2016googlectc,audhkhasi2017a2w,audhkhasi2018a2w,li2018advancing,yu2018multistage} jointly model the acoustic, pronunciation, and language models at the word level under a unified framework. Word-level modeling avoids the need for additional decoding, but introduces new challenges. By modeling acoustics at the word level, the system needs to deal with significant variability in word duration, and it is challenging to learn to recognize infrequent words.
A2W systems perform well given access to very large amounts of training data. For example, ~\cite{soltau2016googlectc} trained a 100K-word vocabulary A2W model and matched the performance of a state-of-the-art sub-word CTC model, but required 125K hours of training speech. Other recent work~\cite{audhkhasi2017a2w, audhkhasi2018a2w, yu2018multistage} has explored techniques for training on much less data (e.g., 300 hours), but performance gaps still remain between A2W and sub-word models.

In this work we develop techniques for addressing the challenge of infrequent words in A2W recognition.  In an end-to-end neural A2W model, the final weight layer consists of one vector per word in the vocabulary, which can be seen as a word embedding matrix.  Most weights in this large matrix are associated with very few training examples since most words are rare.  In addition, out-of-vocabulary words cannot be predicted at all (unlike in sub-word models).

Our approach is to first learn a word embedding model in a data-efficient way, and then to use it in two ways:  (1) By training the recognizer so as to retain similar weights to the pre-trained embeddings, and (2) for predicting words that are unseen in training.  The embedding model learns shared structure between words, and therefore generalizes well to rare or unseen words.  Our pre-trained embedding model is learned using techniques from recent work on acoustic word embeddings, which has found that high-quality discriminative embeddings can be learned from very little data (e.g., $\sim$100 minutes)~\cite{Kamper_16a,settle2016discriminative,he+etal_iclr2017,settle2017query}.  In particular, our embedding approach closely follows that of~\cite{he+etal_iclr2017}, which jointly learns an acoustic embedding function---mapping an acoustic signal to a fixed-dimensional vector---and an {\it acoustically grounded} textual embedding function---mapping a character sequence to a vector.  

We explore a number of variants of these ideas, and find that A2W recognition performance is consistently improved by either initializing the recognizer's word embeddings with acoustically grounded embeddings or by regularizing toward them.  We also introduce a simple method for predicting words outside of the training vocabulary, which improves performance when the training vocabulary is limited.

%% file: approach.tex
\vspace{-.1in}
\section{Approach}
\label{sec:approach}
\vspace{-.05in}

\subsection{Acoustics-to-Word Model}
\label{ssec:approach-a2w}
\vspace{-.05in}
The acoustics-to-word (A2W) model uses a single recurrent neural network, typically a bidirectional long short-term memory network (BLSTM), trained with the connectionist temporal classification (CTC) loss to recognize words from input acoustic sequences. Prior work has found that A2W models either require very large amounts of training data~\cite{soltau2016googlectc} or careful training when using limited amounts of training data~\cite{audhkhasi2017a2w,audhkhasi2018a2w}. In particular,~\cite{audhkhasi2018a2w} showed that presenting utterances in increasing order of length, initializing with a phone CTC BLSTM, and dropout contributed to significant improvements in the word error rate (WER). This recipe, when applied to the (intermediate sized) 2000-hour Switchboard-Fisher training set, produced a WER on par with several state-of-the-art sub-word based models at the time.

Despite some success training competitive A2W models with large amounts of data, they lag behind conventional models when given more limited training data. Furthermore, an A2W model is trained with a fixed vocabulary and cannot recognize out-of-vocabulary (OOV) words. Several prior approaches have been developed to improve the OOV recognition performance of an A2W model. This includes the spell-and-recognize model~\cite{audhkhasi2018a2w} that is trained to predict the character sequence of a word followed by the the word itself. This enables the model to backoff from an unknown word token to its sequence of characters. Another approach is to train a multi-task CTC network to predict both word and character sequences~\cite{li2017acoustic}.

\begin{figure}
\centering
\includegraphics[width=0.9\linewidth]{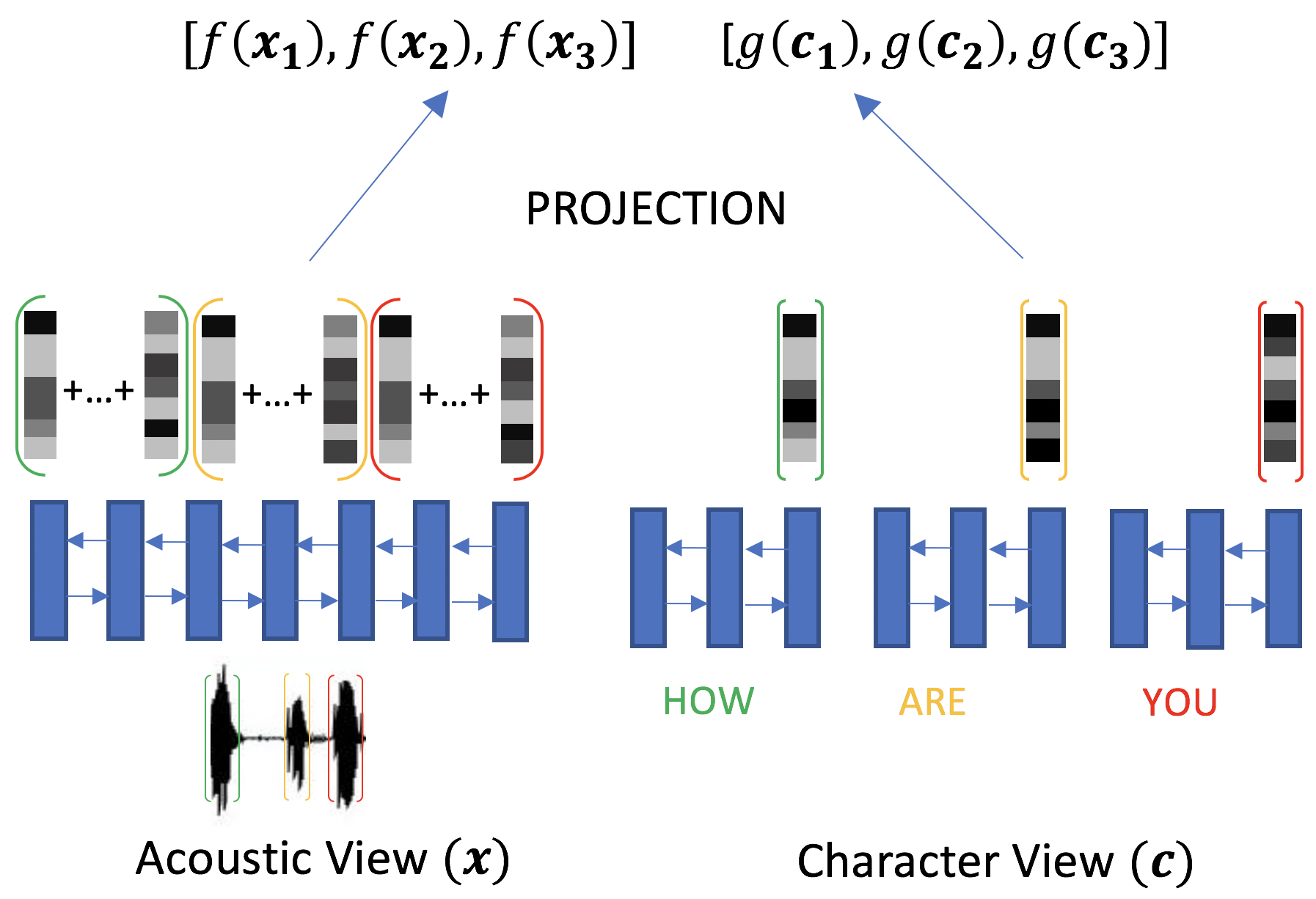}
\caption{Acoustically grounded word embeddings from an utterance-transcript pair.  The utterance $x$ is fed through the Acoustic View BLSTM, and the hidden state outputs are averaged over each word's alignment region.  The character sequence $c$ is fed through the Character View BLSTM and the final output hidden state is retained.  Finally, the outputs from each view are passed through a projection layer.}
\vspace{-0.125in}
\label{fig:a2w}
\end{figure}

\vspace{-0.075in}
\subsection{Acoustically Grounded Word Embeddings}
\label{ssec:approach-awe}

Word embeddings, or continuous vector representations of words, are a common tool in natural language processing, and are typically used to represent the meaning (semantics) of words~\cite{deerwester1990indexing,bengio2003neural,mnih2007three, mikolov2013distributed,pennington2014glove}.  The final (pre-softmax) layer of weights in an A2W model consists of one vector per word in the vocabulary, and can therefore be viewed as embeddings of those words. In fact, in earlier work on A2W-based speech recognition~\cite{audhkhasi2017a2w}, the final layer was initialized with GloVe word embeddings~\cite{pennington2014glove}.
In this work, we investigate the effect of learning a final weight layer which is encouraged to match externally trained word embeddings. Rather than using semantic word embeddings, we consider whether {\it acoustically grounded} embeddings---that is, embeddings that encode acoustic-phonetic similarity rather than semantic similarity---may be helpful.

Recent work has explored a number of acoustic and acoustically grounded word embedding approaches.  Several approaches have been developed for learning {\it acoustic word embedding} models---functions mapping arbitrary-duration spoken word signals to fixed-dimensional vectors---so as to encode either phonetic~\cite{maas+etal_icmlwrl12,levin+etal_asru13,guoguo+etal_icassp15,Kamper_16a,chung2016unsupervised,settle2016discriminative,audhkhasi2017end} or semantic~\cite{chung+glass_interspeech18} information, or both~\cite{chen2018phonetic}.

Other work has considered {\it acoustically grounded word embeddings}, that is embeddings of written words that encode their acoustic/phonetic content~\cite{BengioHeigol14a,ghannay2016evaluation,audhkhasi2017end,he+etal_iclr2017}.\footnote{The term {\it acoustic word embedding} is sometimes used to refer to embeddings of either spoken or written words.  To clarify the distinction, we use {\it acoustically grounded word embedding} for embeddings of written words.}  
Our approach, sketched in Figure~\ref{fig:a2w}, is based on that of~\cite{he+etal_iclr2017}, where two embedding functions are learned jointly, one for acoustic signals (spoken words) and one for character sequences (written words).

We learn two embedding models, $f$ and $g$, which map acoustic sequences ${\bf x}$ and character sequences ${\bf c}$, respectively, to fixed-dimensional vectors.
The acoustic embedding model consists of a stacked BLSTM followed by a sum over the output layer hidden states and a projection to a lower-dimensional vector in $\Re^d$, which is the acoustic embedding $f({\bf x})$. The character sequence embedding model consists of a learned character embedding layer and a single-layer BLSTM; the final hidden state is projected to $\Re^d$ and the result is used as the (acoustically grounded) textual embedding vector $g({\bf c})$. Here we use a shared final projection layer.

We learn the embedding functions $f$ and $g$ jointly so that same-word pairs are mapped to similar vectors while different-word pairs are mapped far apart. Let $d$ denote the cosine distance, $d(x, y) = 1 - \frac{x \cdot y}{\Vert x \Vert \Vert y \Vert}$, $m$ a margin hyperparameter, and $char({\bf x})$ the character sequence corresponding to the word label of acoustic sequence ${\bf x}$.  We learn using a sum of two multi-view objectives (namely objectives $0$ and $2$ of~\cite{he+etal_iclr2017}):

\vspace{-0.5cm}
\begin{flalign}
\min_{f,g} &\sum_{i=1}^N \left[ m + d(f({\bf x}_i), g({\bf c}_i)) - \min_{{\bf c} \ne {\bf c}_i} d(f({\bf x}_i), g({\bf c})) \right]_{+}&\notag\\
	&+\sum_{i=1}^N \left[ m + d(g({\bf c}_i), f({\bf x}_i)) - \min_{char({\bf x}) \ne {\bf c}_i} d(g({\bf c}_i), f({\bf x}))\right]_{+}
\label{eq:mv} 
\end{flalign}
\vspace{-0.4cm}

\noindent where $N$ is the number of training pairs $({\bf x}_i, {\bf c}_i)$.  In practice, we do not minimize over all ${\bf c} \ne {\bf c}_i$ and all $char({\bf x}) \ne {\bf c}_i$, but rather select the $k$ most offending examples within each mini-batch and use the mean of their cosine distances~\cite{schroff2015facenet}.

Finally, we use the acoustic segment embedding function $f$ and the character sequence embedding function $g$ in pretraining the A2W speech recognizer. Notice that, since $g$ can be applied to arbitrary character sequences, it is in principle applicable to words that have been seen very few times, or not at all, in training.

\vspace{-0.05in}
\subsection{Acoustically Grounded Word Embeddings for Recognition}
\label{ssec:approach-integration1}

In prior work on A2W modeling~\cite{audhkhasi2017a2w,audhkhasi2018a2w,yu2018multistage}, careful model initialization and regularization techniques are cited as essential for effective training on limited data. Two techniques explored in prior work were phone CTC pretraining and GloVe embedding initialization at the word-level prediction layer. Without such initializations, early training methods in this area failed to converge at all~\cite{audhkhasi2017a2w}. Since GloVe embeddings are trained such that proximity in the embedding space implies similarity in semantics, we may be able to improve performance by instead utilizing an embedding space optimized for acoustic-phonetic similarity.
In particular we propose using the acoustically grounded embeddings trained with the contrastive loss of Eq.~\ref{eq:mv}.

We consider using our embeddings in several ways for improved training of the prediction layer weights:  (1) {\bf initializing} the weights with the word embeddings, and then training as usual; (2) {\bf regularizing} the weights to remain similar to the word embeddings, after initializing as in (1); and (3) {\bf freezing} the weights at the initialized values given by the word embeddings.

For the regularization approach, we train the recognizer with a training objective that is a weighted average of the baseline recognizer loss and an embedding regularization L2 loss:

\vspace{-0.5cm}
\begin{flalign}
\mathcal{L}({\bf X}, {\bf Y}) &= (1-\lambda) \sum_{i=1}^N \mathcal{L}_{CTC} ({\bf x}_i,{\bf y}_i) +\notag \\
     & \lambda \sum_{y \in \displaystyle \cup_{i=1}^N {\bf y}_i} \Vert g(char(y))-w(y) \Vert^2
\label{eq:loss}
\end{flalign}
\vspace{-0.25cm}

\noindent where ${\bf X} = \{{\bf x}_i\}_{i=1}^N$ and ${\bf Y} = \{{\bf y}_i\}_{i=1}^N$ are a batch of $N$ utterance-transcript pairs, $y$ is a word and char($y$) is its character sequence, $g$ is the character sequence embedding function, $w(y)$ is the row of the CTC prediction layer for word $y$, and $\mathcal{L}_{CTC}$ is the usual CTC loss used for the baseline recognizer.

The last approach, of freezing the prediction layer after initialization, is the extreme case of optimizing for CTC while strictly adhering to the embedding initialization.  One helpful feature of this approach is that it can naturally apply to an extended vocabulary at decoding time, some of which was unseen at training time. When using the model to decode, we first predict outputs from the training vocabulary; however, if we predict an $<$UNK$>$ token, then we rescore by removing $<$UNK$>$ and predicting from the entire extended vocabulary. Early work on learning acoustic word embeddings~\cite{maas+etal_icmlwrl12,BengioHeigol14a} used word-level acoustic representations precisely for rescoring in large-vocabulary speech recognition systems. In this work, we show the ability for our learned embedding function $g$ to accurately extend to arbitrary unseen words without any additional training.

%% file: experiments.tex
\vspace{-0.1cm}

\section{Experimental Setup}
\label{sec:setup}

\vspace{-.075in}
\subsection{Data}
\label{ssec:data}
We use the standard 300-hour Switchboard corpus of conversational English speech. Speaker-independent 40-dimensional log-Mel features are computed with the addition of $\Delta$s+$\Delta\Delta$s and stacking+frame skipping with a rate of 2, resulting in 240-dimensional input acoustic features. Vocabularies of approximately 4K, 10K, and 20K words are used, corresponding to minimum occurrence thresholds set at 25, 6, and 2, respectively. Experiments on out-of-vocabulary extension use a 34K vocabulary for rescoring.

For training the acoustic word embeddings, character sequences are composed of symbols from a 35-character vocabulary: 26 English letters, 6 punctuation symbols ({[]}$<>$-'), and 3 sounds ({[NOISE]}, {[VOCALIZED-NOISE]},{[LAUGHTER]}). Words containing digits are spelled out using this vocabulary (e.g. ``7-11" is spelled ``SEVEN-ELEVEN"). Unspoken parts of partial words are in {[]} with partial starts and ends denoted by $<$ and $>$, respectively (e.g. ``$<$[YO]UR" and ``YO[UR]$>$"). Acoustic word segment boundaries are obtained from word-level forced alignments produced by a competitive ASR system for Switchboard. When conducting word embedding training, word segments shorter than 6 frames and words outside the training vocabulary are omitted from the embedding loss computation. During pre-processing, if these restrictions omit all words from an utterance, then the utterance itself is removed. The same set of utterances is then used for training both the word embeddings and the recognizer.

\vspace{-.1in}
\subsection{Experiments}
\label{ssec:expts}

During training of both the word embedding and recognition models, a batch size of 64 utterances is used, split across 4 GPUs. The same learning rate reduction scheme is used for both:  If the held-out performance (see Section ~\ref{sssec:awe} and~\ref{sssec:a2w}) fails to improve after 4 epochs, the learning rate is decayed by a factor of 10 and the model is reset to the previous best. The acoustically grounded word embedding model is trained using the Adam optimizer~\cite{kingma2015adam} with an initial learning rate of $0.0005$ and parameters $\beta_1 = 0.9$, $\beta_2 = 0.999$, and $\epsilon = 10^{-8}$.
The A2W recognition model is trained using stochastic gradient descent (SGD) with Nesterov momentum~\cite{nesterov1983method} with an initial learning rate of $0.02$ and momentum of $0.9$. Training is stopped when the held-out performance stops improving and the learning rate is $< 10^{-8}$. All experiments were conducted using the PyTorch toolkit~\cite{paszke2017automatic}.

\vspace{-.075in}
\subsubsection{Acoustically grounded word embeddings}
\label{sssec:awe}

The acoustic view model is composed of a 6-layer BLSTM with 512 hidden units (per direction per layer), which is initialized with a BLSTM trained as a phone CTC recognizer. The acoustic BLSTM takes full utterances as input, but to produce a word embedding for a given word segment within an utterance, the hidden state outputs corresponding to frames in that segment are averaged to produce a single 1024-dimensional vector. The character view model includes an initial character embedding layer that maps each of the 35 characters to a 64-dimensional vector, followed by a 1-layer BLSTM with 512 hidden units (per direction). A projection layer is used to transform the 1024-dimensional vectors output by the acoustic and character view models to 256-dimensional embeddings. Unlike CTC, the loss is not calculated for each utterance in isolation. Since the multi-view objective (Equation~\ref{eq:mv}) involves sampling the $k$ most offending examples with respect to each acoustic segment and each character sequence, distributing across 4 GPUs limits samples to those present in only 16 utterances. However, we find that distributing speeds up training without degradation in performance. We start with $k=15$ and reduce to $k=5$ over the first 300 mini-batches. 

As in~\cite{he+etal_iclr2017}, we evaluate the quality of our embeddings using a cross-view word discrimination task applied to the held-out set. Given a word and an acoustic segment, the goal is to determine if the acoustic segment is a spoken instance of the word. If the cosine distance between their embeddings is below a threshold, we consider them a match. We evaluate performance via the average precision (AP) over all thresholds. This AP is our held-out performance measure, used to tune the embedding model before using it in the A2W recognizer. For $M$ acoustic word segments and a vocabulary of $N$ words, we compute the average precision over $M \times N$ pairs.
\vspace{-0.225cm}

\subsubsection{Acoustics-to-word recognition model}
\label{sssec:a2w}

The A2W model is composed of a 6-layer BLSTM with 512 hidden units (per direction per layer), a 256-dimensional linear projection layer, and a prediction layer with dimension determined by the vocabulary size. Dropout ($p=0.25$) is used between BLSTM layers. The baseline A2W system uses a phone CTC initialization for the BLSTM, while feed-forward layers are randomly initialized as in~\cite{audhkhasi2018a2w}. During A2W training, the held-out WER is used to measure performance.

\vspace{-0.05cm}

\begin{table}[!htbp]
\centering
    \begin{tabular}{@{}lc@{}}
        \toprule
        \multicolumn{1}{c}{Initialization} &
        \multicolumn{1}{c}{SWB Dev WER (\%)}\\
        \midrule
		Phone CTC init & 17.5\%\\
		Phone CTC init + AWEs & 17.6\%\\
		Acoustic view init + AWEs & 16.7\%\\
        \bottomrule
    \end{tabular}
\caption{SWB development (held-out) word error rates for different methods of initializing the A2W model, using a 10k-word vocabulary.}
\label{tab:initialize}
\end{table}
\vspace{-0.25cm}

For word embedding integration experiments, the A2W model is initialized with the acoustic view BLSTM and the 256-dimensional projection layer from multi-view training. The prediction layer is initialized with unit-normalized word embeddings output by the character view model for each word in the vocabulary (with the exception of $<$BLANK$>$ and $<$UNK$>$, which are randomly initialized). We find that multi-view pre-training of the BLSTM and projection layer is essential to improvement over the baseline system (see Table~\ref{tab:initialize}).

Regularization experiments penalize deviation (L2 distance) of the prediction layer weights from the word embedding initialization. This penalty is only applied with respect to those words present in a given mini-batch. The hyperparameter $\lambda \in \{0.1, 0.25, 0.5, 0.75, 0.9, 0.99\}$ is tuned to manage the trade-off between the CTC objective and this penalty (Equation~\ref{eq:loss}).

At the extreme end of regularization, we experiment with freezing the prediction layer after initialization, including the randomly initialized $<$BLANK$>$ and $<$UNK$>$ tokens. By retaining the
consistency of the learned embedding space throughout training, we can acquire new embeddings for any OOV words by running their character sequences through the character view model. We conduct experiments with OOV prediction by concatenating these new word embeddings to the prediction layer, and rescoring with the extended vocabulary whenever an $<$UNK$>$ is predicted.

%% file: results.tex
\vspace{0.5cm}
\begin{table}[!htbp]
\centering
    \begin{tabular}{@{}ccccc@{}}
        \toprule
        \multicolumn{1}{c}{Vocab} &
        \multicolumn{1}{c}{AGWE} &
        \multicolumn{3}{c}{CTC}
        \\
        \cmidrule{3-5}	& 		& Baseline	& Initialized	& Regularized\\\midrule
		4K  			& 0.894	& 0.489		& 0.719 			& 0.762 		\\
		10K 			& 0.879	& 0.279		& 0.644			& 0.734		\\
		20K 			& 0.858	& 0.160 		& 0.633			& 0.596		\\
        \bottomrule
    \end{tabular}
    \caption{Cross-view word discrimination performance, measured via average precision (AP), of acoustically grounded word embeddings (AGWE) and CTC-based embeddings (prediction layer weights).}
\label{tab:ap}
\end{table}

\vspace{-0.35cm}
\section{Results}
\label{ssec:results}
\vspace{-0.05in}
In Table~\ref{tab:ap}, we compare the quality of acoustically grounded word embeddings (AGWE) trained explicitly using the multi-view objective against CTC-based embeddings given by the prediction-layer weights after CTC training.  The significantly better AP of AGWE shows that they capture discriminative information that is not discovered implicitly by CTC training. By using these pre-trained embeddings to initialize our model or regularize CTC training, the prediction layer is better able to retain this word discrimination ability.
\vspace{-0.05cm}

\begin{table}[!htbp]
\centering
    \begin{tabular}{@{}lccc@{}}
        \toprule
        \multicolumn{1}{c}{System} &
        \multicolumn{3}{c}{Vocab}
        \\
        \cmidrule{2-4}	& $4$K 				& $10$K 				& $20$K\\\midrule
		Baseline		& 16.4/25.7 			& 14.8/24.9 			& 14.7/24.3\\
		Initialized 	& 15.6/{\bf 25.3} 	& 14.2/{\bf 24.2} 	& 13.8/24.0\\
		Regularized 	& 15.5/25.4			& {\bf 14.0}/24.5	& {\bf 13.7/23.8}\\
		Frozen			& 15.6/25.6			& 14.6/24.7			& 14.2/24.7\\
		\indent+OOV rescoring	& {\bf 15.0/25.3} 	& 14.4/24.5			& 14.2/24.7\\
        \midrule
        Curriculum~\cite{yu2018multistage} & - & - & 13.4/24.2\\
        Curriculum+Joint CTC/CE~\cite{yu2018multistage} & - & - & 13.0/23.4\\
        \bottomrule
    \end{tabular}
    \caption{Results (\% WER) on the SWB/CH evaluation sets.  The best result for each data set and each vocabulary size is boldfaced.}
\label{tab:eval}
\end{table}
\vspace{-0.05cm}

Table~\ref{tab:eval} shows evaluation results on the Switchboard (SWB) and CallHome (CH) test sets. We find that both embedding initialization and regularization improve over the baseline WERs.  For the regularized model, the values of $\lambda$ (tuned on held-out data) are $0.25$, $0.25$, and $0.5$ for the $4$K, $10$K, and $20$K vocabularies.  However, all values of $\lambda$ yield improvements on the held-out set over the baseline.

Although outperformed by the embedding-initialized and regularized models, the model trained with a frozen prediction layer (``Frozen'') also consistently outperforms the baseline, with the exception of the $20$K CH evaluation.  An added feature of the Frozen model, as discussed in Section~\ref{ssec:approach-integration1}, is that it allows for straightforward OOV extension via rescoring. Table~\ref{tab:eval} shows that vocabulary extension and rescoring using the Frozen model improves performance considerably when using the smallest vocabulary.  For the $4$K vocabulary recognizer, this approach results in an overall absolute WER reduction of $1.4\%$ over the baseline model. We also note that the relative improvement seen by adding OOV rescoring to the $10$K vocabulary Frozen model is similar to that offered by the spell-and-recognize system in~\cite{audhkhasi2018a2w} without the need for additional training.

Recent work~\cite{yu2018multistage} shows strong results using a multi-stage A2W approach including curriculum learning from the $10$K to the $20$K vocabulary, joint CTC/cross entropy (CE) training, and data augmentation. In Table~\ref{tab:eval}, we report results from their most comparable setups, curriculum and joint CTC/CE~\cite{yu2018multistage}. Future work may improve further upon these results by combining embedding regularization with the techniques from~\cite{yu2018multistage}.

Inspecting outputs from the Frozen+OOV rescoring model, we find that when the A2W system produces an $<$UNK$>$ prediction in place of a single word, we often accurately recover the correct word within the
top hypotheses, as seen in the first two rows of Table~\ref{tab:analysis}. The majority of remaining mistakes correspond to the first-pass model predicting $<$UNK$>$ in place of multiple words or part of a word.
In such cases we cannot recover the correct word, but we find that many predictions are reasonable phonetic matches for the ground truth.
For example, in the third example in Table~\ref{tab:analysis}, the first-pass model combines two words ``LOANS ARE'' into a single $<$UNK$>$, and the rescoring model produces a close phonetic match, ``LOANER''.
Other typical examples include splitting up compound words such as ``CAREGIVER" and words that are outside the extended 34K vocabulary such as ``CANTEENS".

\vspace{-0.125cm}

\begin{table}[!htbp]
\footnotesize\centering
    \begin{tabular}{@{}l@{}}
        \toprule
        \textbf{REF:} some REMINDERS for me as we are talking\\
        \textbf{HYP (1st pass):} some $<$UNK$>$ for me as we are talking\\
        \textbf{HYP (rescoring):} some \textbf{REMINDERS} for me as we are talking\\
        \midrule
        \textbf{REF:}   fair and speedy TRIAL\\
        \textbf{HYP (1st pass):}   fair and speedy $<$UNK$>$\\
        \textbf{HYP (rescoring):}   fair and speedy \textbf{TRIAL}\\
        \midrule
		\textbf{REF:} but those LOANS ARE so much cheaper\\
		\textbf{HYP (1st pass):} 	but those $<$UNK$>$ so much cheaper\\
        		\textbf{HYP (rescoring):} but those \textbf{LOANER} so much cheaper\\
		\midrule
		\textbf{REF:} one particular CAREGIVER AND then that one\\
        \textbf{HYP (1st pass):} one particular CARE $<$UNK$>$ then that one\\
		\textbf{HYP (rescoring):} one particular CARE \textbf{GIVER} then that one\\
		\midrule
		\textbf{REF:} bring two CANTEENS  just to make sure\\
        \textbf{HYP (1st pass):} bring two $<$UNK$>$ just to make sure\\
		\textbf{HYP (rescoring):} bring two \textbf{CAMPING'S} just to make sure\\
        \bottomrule
    \end{tabular}
	\caption{Successes and failures in OOV prediction with the Frozen+OOV rescoring model.}
\label{tab:analysis}
\end{table}

%% file: conclusion.tex
\vspace{-0.35cm}
\section{Conclusion}
\label{sec:conc}
\vspace{-0.05in}

We have introduced techniques for using pre-trained acoustically grounded word embeddings for improving acoustics-to-word CTC speech recognition models.  We have found that consistent performance improvements can be obtained by incorporating embeddings through initialization, regularization, and out-of-vocabulary prediction.  For example, by regularizing the recognizer prediction layer toward the embeddings, we obtain $0.8-1\%$/$0.4-0.5\%$ absolute WER improvements for the Switchboard/CallHome data sets.  If we also rescore with an expanded vocabulary to resolve OOVs, then in the small-vocabulary (4k-word) case we can improve the WER by a total of $1.4\%$ absolute on Switchboard.

Future directions include exploring additional kinds of embedding models and training criteria, as well as tighter integration of the embedding and recognizer training.  Another promising direction, considering the encouraging results with small vocabulary sizes, is to apply these ideas to the recognition of low-resource languages.